\documentclass[10pt, a4paper]{article}
\usepackage{lrec}
\usepackage{multibib}
\newcites{languageresource}{Language Resources}
\usepackage{graphicx}
\usepackage{tabularx}
\usepackage{soul}

\usepackage{epstopdf}
\usepackage[utf8]{inputenc}

\usepackage{hyperref}
\usepackage{xstring}

\usepackage{color}

\usepackage{pgfplots}
\pgfplotsset{compat=1.18}
\usetikzlibrary{patterns}
\usetikzlibrary{intersections}
\usepgfplotslibrary{fillbetween}
\usepackage{booktabs}
\usepackage{dblfloatfix}
\usepackage{amsmath}

\newcommand{\corpusname}{Rezwan}

\title{\corpusname: Leveraging Large Language Models for Comprehensive Hadith Text Processing: A 1.2M Corpus Development}

\name{\\
\textbf{Majid Asgari-Bidhendi$^{2,1}$, Muhammad Amin Ghaseminia$^{2,1}$, Alireza Shahbazi$^{1}$,}\\
\textbf{Sayyed Ali Hossayni$^{1}$, Najmeh Torabian$^{3,1}$, Behrouz Minaei-Bidgoli$^{2}$}
}

\address{
$^{1}$ Noor Avaran Jelvehaye Maanaei Najm Co. \\
$^{2}$ Iran University of Science and Technology
$^{3}$ Islamic Azad University \\
        Tehran, Iran \\
        \{majid, ma.ghaseminia, arsh, sayyed, trbn\}@najm.ac, b\_minaei@iust.ac.ir
}

\abstract{
This paper presents the development of \textit{\corpusname}, a large-scale
AI-assisted Hadith corpus comprising over 1.2M narrations, extracted and
structured through a fully automated pipeline. Building on digital repositories such as \textit{Maktabat Ahl al-Bayt}, the pipeline employs Large Language Models (LLMs) for segmentation, chain--text separation, validation, and multi-layer enrichment. Each narration is enhanced with machine translation into twelve languages, intelligent diacritization, abstractive summarization, thematic tagging, and cross-text semantic analysis. This multi-step process transforms raw text into a richly annotated research-ready infrastructure for digital humanities and Islamic studies. A rigorous evaluation was conducted on 1,213 randomly sampled narrations, assessed by six domain experts. Results show near-human accuracy in structured tasks such as chain--text separation (9.33/10) and summarization (9.33/10), while highlighting ongoing challenges in diacritization and semantic similarity detection. Comparative analysis against the manually curated Noor Corpus demonstrates the superiority of Najm in both scale and quality, with a mean overall score of 8.46/10 versus 3.66/10. Furthermore, cost analysis confirms the economic feasibility of the AI approach: tasks requiring over 229,000 hours of expert labor were completed within months at a fraction of the cost. The work introduces a new paradigm in religious text processing by showing how AI can augment human expertise, enabling large-scale, multilingual, and semantically enriched access to Islamic heritage.
\\ \newline \Keywords{Hadith corpus, Large Language Models, Digital Humanities, Islamic NLP, Semantic Enrichment} }

\begin{document}

\maketitleabstract

\section{Introduction}

The hadith tradition, comprising the sayings of the Prophet Muhammad (pbuh) and the words of the Imams (as), represents the second most authoritative source of Islamic knowledge after the Qur'an. It has profoundly influenced the development of Islamic sciences such as jurisprudence, theology, exegesis, and ethics. A hadith is composed of two main parts: the \textit{chain} of narrators (also known as \textit{isnad}), and the main \textit{text} of the narration (also known as \textit{matn}). However, the vast scale, scattered nature of the sources, and the linguistic and structural complexity of these texts have always posed significant challenges to researchers. Traditional approaches to collecting, curating, and analyzing hadith are often manual, labor-intensive, and heavily dependent on specialized expertise, which makes large-scale research extremely difficult.

With the rise of the digital era and the remarkable advances in Artificial Intelligence (AI), particularly the development of Large Language Models (LLMs), new opportunities have emerged for overcoming these long-standing challenges. These transformative technologies enable the automation of complex tasks that were previously exclusive to human specialists.

In this paper, we present a novel and systematic approach for building and enriching a large-scale hadith corpus using AI. We describe an end-to-end automated processing pipeline that not only performs the fundamental tasks of extraction and structuring of hadith texts, but also provides multiple enrichment layers, including machine translation into 12 languages, intelligent diacritization, summarization, thematic tagging, and the discovery of intricate semantic and lexical relationships across narrations.

The aim of this work is twofold: first, to demonstrate the feasibility of constructing a massive and richly annotated hadith corpus that can serve as a powerful research tool; and second, to provide a rigorous evaluation of the quality and economic value of such an AI-driven approach. By doing so, this study contributes to the field of Digital Humanities and Islamic Studies, offering a new paradigm for large-scale textual analysis in religious scholarship.

\section{Related Work}

Natural language processing of religious and historical corpora has attracted growing interest. For example, surveys of Qur'anic NLP note extensive work on tasks from morphology to semantic search~\cite{bashir2022computational}. Bashir et al. (2022) emphasize that hadith literature is a vast complementary source to the Qur'an, suggesting that neglecting it can yield incomplete insights~\cite{bashir2022computational}. Arabic morphological analyzers and corpora have been developed for classical texts (e.g., Qur'an, Bible, classical Arabic) to support downstream tasks~\cite{bashir2022computational}.

Recent work on summarization, semantic Q\&A, and entity linking in religious texts underscores this trend. For instance, Alqarni (2024) demonstrates that large LLM embeddings can improve semantic search over Qur'anic text~\cite{alqarni2024embedding}, and Khair and Sawalha (2025) use GPT-4 to translate 250K pages of Arabic Islamic literature (Qur'an, Hadith, juristic works) into English, validating translation quality by back-translation~\cite{khair2025automated}. Other efforts use retrieval-augmented generation for religious Q\&A: Alan et al. (2025) build ``MufassirQAS,'' a Turkish-language Islamic chatbot that grounds answers in a verified corpus of tafsir and hadith, thereby reducing LLM hallucinations and citing source passages~\cite{alan2025mufassir}. These studies illustrate how LLMs and knowledge-graph techniques (e.g. SemanticHadith~\cite{kamran2024semantic}) can enrich religious text processing with advanced NLP.

Efforts to create Hadith corpora and digital libraries remain limited but are emerging. Altammami et al. (2019) introduced the Leeds--King Saud University (LK) corpus: a bilingual Arabic--English parallel corpus of 39,038 authenticated hadiths from the six major collections. They used a domain-specific segmentation tool to annotate each hadith's chain and text~\cite{altammami2019arabic}. Alosaimy and Atwell (2017) earlier built the Sunnah Arabic Corpus~\cite{alosaimy2017sunnah}, a smaller annotated corpus ($\approx$144K words) extracted from Riyad al-Salihin~\cite{altammami2019arabic}. More recently, Taghreed Tarmom et al. (2022) released a corpus of non-authentic (fabricated) hadiths and demonstrated transformer-based detection, while noting the concurrent release of the LK hadith corpus~\cite{tarmom2022non}. Surveys by Azmi et al. (2019) and others highlight that computational hadith research is ``in its infancy,'' with few reusable datasets available and many studies relying on private data~\cite{altammami2019arabic}. Zaghouani (2017) similarly reports that a known Arabic hadith corpus had become inaccessible, underscoring the need for shared resources. In addition to corpora, several projects structure hadith data via ontologies and knowledge graphs. For example, the SemanticHadith project (Kamran et al., under review) develops an ontology for hadith concepts and builds a knowledge graph from six collections, extracting entities and relations with NLP techniques to enable semantic search~\cite{kamran2024semantic}. Altharwa et al. (2024) explore topic modeling in hadith (``RoBERT2Vec'') to identify salient themes in the text~\cite{aftar2024robert2vectm}. Together, these efforts form the foundation of Islamic digital resources, but none match the scale of our 2.3M-hadith pipeline.

Applications of LLMs for knowledge extraction and translation in religious domains are also expanding. The emerging field of Islamic NLP includes machine translation of scriptural texts and semantic querying. For instance, Azmi et al. note that Qur'anic text search can be enhanced by leveraging hadith references~\cite{bashir2022computational}, and several works build Qur'an--hadith ontologies to enable concept-level search~\cite{bashir2022computational}. In machine translation, Khair \& Sawalha (2025) used GPT-4 to automate translation of Islamic literature, demonstrating that LLMs can provide high-quality, low-cost translations of sacred texts~\cite{khair2025automated}. Semantic search tools for the Qur'an have been developed as well: Alqahtani and Atwell (2016) built an ontology-based semantic search for Arabic Qur'anic queries~\cite{bashir2022computational}, and Alqarni (2024) shows that LLM embeddings outperform traditional methods at higher-level semantic queries of the Qur'an~\cite{alqarni2024embedding}. Question-answering systems tailored to hadith have also been proposed: Abdi et al. (2019) describe a hadith QA system that combines semantic similarity measures and query expansion to improve answer relevance~\cite{abdi2019question}. Retrieval-augmented LLMs, like MufassirQAS~\cite{alan2025mufassir}, further illustrate how vector search over indexed religious texts can ground and validate answers.

Finally, our work relates to multi-layer text annotation pipelines in digital humanities. Precedents include systems that integrate segmentation, entity extraction, and linking across documents. Alberani et al. (2019) present a pipeline for unstructured hadith and narrator biography texts: they segment hadiths, tag morphology/POStags, then construct a graph of narrators across documents to boost entity extraction~\cite{alberani2019arabic}. Their cross-document approach improved extraction of narrators and relations by leveraging an induced narrator graph. In the broader NLP context, recent summarization systems use layered pipelines as well: for example, Essa et al. (2025) integrate Arabic named-entity recognition into an abstractive summarization model, significantly improving summary coherence~\cite{essa2025enhanced}. Ontology-driven annotation pipelines have been applied to sacred texts too: linked-data efforts (e.g., SemanticHadith~\cite{kamran2024semantic}) use NER and ontological tagging to semantically enrich hadith text. These multi-step frameworks -- involving segmentation, linguistic annotation, translation and linking -- inform our design. Our automated LLM-based pipeline builds on this tradition by unifying segmentation, validation, multi-language translation, diacritization, summarization, semantic analysis, and inter-hadith linking into one scalable process. In sum, while prior work has applied NLP at various levels to Islamic texts (from corpus creation~\cite{altammami2019arabic} to QA and summarization~\cite{abdi2019question,essa2025enhanced}), this paper represents a novel large-scale integration of these techniques for hadith.

\section{Corpus Construction Methodology}

The construction of the proposed hadith corpus was carried out through a fully automated pipeline, carefully designed to address the unique challenges of processing large-scale religious texts. This section describes the methodology in three major components: data collection and preprocessing, pipeline architecture, and enrichment through annotation.

\subsection{Data Collection and Preprocessing}

The foundation of any corpus lies in the quality and coverage of its raw data. For this project, several well-known digital repositories of hadith and related Islamic texts were evaluated, including \textit{OpenITI}, \textit{al-Maktaba al-Shamela}, and \textit{IslamWeb}. After comparative analysis, the \textit{Maktabat Ahl al-Bayt (AS)} collection was selected as the primary source due to two main reasons:

\begin{itemize}
    \item \textbf{Comprehensiveness:} The collection provides extensive coverage of both Sunni and Shia sources, fulfilling the requirement of building a unified and representative corpus.
    \item \textbf{Structural consistency:} The files are organized in a relatively coherent format, which facilitates automated extraction and processing.
\end{itemize}

Despite these advantages, significant preprocessing was required. Many hadith works were misclassified under categories such as ``Fiqh'' or ``Tafsir'' instead of ``Hadith.'' Consequently, a manual filtering and reclassification step was performed to ensure that all texts containing narrations were correctly identified and included in the corpus. This meticulous curation ensured maximum coverage of authentic hadith sources before entering the automated pipeline.

\subsection{Pipeline Architecture}

At the core of the methodology lies an intelligent, multi-stage pipeline designed to transform raw text into a richly structured corpus. The architecture includes the following stages:

\begin{enumerate}
    \item \textbf{Segmentation and Hadith Boundary Detection:}
    Identifying the start and end of narrations is a critical step. Given the variability of hadith structures, simple rule-based methods were insufficient. Instead, prompt-engineered Large Language Models (LLMs) were employed to separate \textit{chain} (chain of transmission) from \textit{text} (main text). A dynamic mechanism based on ``semantic units'' was introduced to avoid truncation of long narrations and to recover incomplete segments.

    \item \textbf{Validation and Alignment with Source Texts:}
    Extracted narrations were validated against the original sources to ensure fidelity. Fuzzy string matching techniques were used to handle discrepancies caused by OCR errors or minor textual variations. Each narration was indexed with page and source references for traceability.

    \item \textbf{Automated Enrichment Pipeline:}
    After validation, narrations were passed into multiple enrichment modules powered by LLMs, each adding an analytical or linguistic layer to the corpus.
\end{enumerate}

This modular design allowed for both scalability and adaptability, enabling continuous integration of new data and corrections.

\subsection{Annotation and Enrichment}

The enrichment process represents one of the most innovative aspects of the corpus. Each narration was enriched with multiple layers of metadata and linguistic annotations, including:

\begin{itemize}
    \item \textbf{Machine Translation:} Every narration was translated into 12 major world languages (e.g., English, Persian, Turkish, Urdu, French, Spanish, German), expanding the accessibility of hadith studies to non-Arabic-speaking audiences.

    \item \textbf{Intelligent Diacritization:} Since more than 80\% of the texts lacked diacritics, an AI-based diacritization model was applied. Its quality was validated against existing diacritization systems, showing superior accuracy.

    \item \textbf{Summarization and Thematic Tagging:} For each narration, concise summaries, key points, and thematic labels were generated, allowing researchers to quickly filter and navigate large sets of narrations.

    \item \textbf{Semantic and Lexical Relationship Discovery:} Using vector-based semantic similarity, narrations were clustered to reveal lexical, semantic, and thematic parallels. This step enabled the discovery of hidden relationships between narrations that transcend keyword-based retrieval.

    \item \textbf{Quality Control Filters:} Automated filters flagged anomalous outputs, such as texts misidentified as hadith, ensuring data reliability.
\end{itemize}

This multi-layered enrichment transformed the corpus from a simple text repository into a dynamic, research-ready infrastructure, facilitating advanced analyses across linguistic, semantic, and comparative dimensions.

\section{Evaluation Framework}

Building a large-scale hadith corpus with advanced AI-based enrichment requires a rigorous and transparent evaluation process to establish its scientific reliability. In this section, we describe the evaluation framework used to assess the structural accuracy, linguistic quality, and analytical depth of the corpus. The framework was designed as a multi-dimensional process combining both quantitative and qualitative measures.

\subsection{Design of the Evaluation Process}

To ensure representativeness, a random sample of 1,213 narrations was drawn from the corpus. The evaluation was conducted by a team of six domain experts in hadith studies. Each expert used a standardized evaluation form that combined numerical scoring (0--10 scale) with qualitative annotations (e.g., marking errors at the word or phrase level). This dual approach allowed for both statistical analysis and error-pattern identification.

The evaluation focused on two complementary dimensions:
\begin{itemize}
    \item \textbf{Quantitative assessment:} Numerical scores for structural accuracy, error rates for tasks like translation and diacritization, and coverage of thematic tags.
    \item \textbf{Qualitative assessment:} Expert evaluation of aspects like the semantic coherence of summaries, the nuanced accuracy of translations, and the relevance of discovered thematic relationships.
\end{itemize}

\subsection{Evaluation Criteria}

The criteria were grouped into three major categories:

\begin{enumerate}
    \item \textbf{Structural and Foundational Accuracy}
    \begin{itemize}
        \item Correct identification of hadith vs. non-hadith texts.
        \item Accuracy of chain–text separation.
        \item Fidelity of extracted narration to its source text.
    \end{itemize}

    \item \textbf{Linguistic and Content Quality}
    \begin{itemize}
        \item Typographical correctness.
        \item Accuracy of diacritization at the character level.
        \item Quality of machine translation (coverage and precision).
        \item Adequacy of summaries and extracted key points.
        \item Accuracy of thematic labels assigned to each narration.
    \end{itemize}

    \item \textbf{Analytical and Relational Accuracy}
    \begin{itemize}
        \item Grouping accuracy of identical narrations across different sources.
        \item Precision in detecting lexical, semantic, and thematic similarity.
        \item Reliability of clustering narrations into coherent groups.
    \end{itemize}
\end{enumerate}

\subsection{Statistical Measures}

For error-based criteria (e.g., diacritization, translation), two well-established measures were applied:

\begin{itemize}
    \item \textbf{Micro-average error rate:} The average error percentage computed per narration, giving equal weight to each narration regardless of its length.
    \item \textbf{Macro-average error rate:} The overall error percentage across all narrations, giving higher weight to longer narrations.
\end{itemize}

This distinction provided a balanced view of performance across both short and long narrations. To avoid distortion by outliers, narrations with more than 60\% error rate in any core dimension (translation, diacritization, or text fidelity) were flagged as ``critical failures'' and excluded from mean score computations, while still being reported separately.

\subsection{Inter-Rater Reliability and Quality Control}

Given the interpretative nature of hadith studies, harmonizing expert judgments was crucial. To achieve consistency:
\begin{itemize}
    \item A shared guideline was iteratively developed through ongoing expert discussions during the evaluation.
    \item Ambiguous cases (e.g., whether a phrase belonged to isnad or matn) were resolved collectively to establish a unified scoring standard.
    \item Error propagation was carefully managed; for instance, a mistranslation leading to flawed summaries was penalized only once, preventing cascading deductions.
\end{itemize}

This process not only improved reliability but also generated a refined guideline that can be reused in future evaluations.

\subsection{Outcome}

The framework ensured that corpus quality was assessed holistically, covering structural correctness, linguistic fidelity, and semantic depth. By combining numerical scoring, error annotation, and expert discussions, the evaluation provided both rigorous statistical evidence and practical insights into strengths and limitations of the AI-driven pipeline.

\section{Results and Analysis}

This section reports the results of the expert-based evaluation of the proposed \textit{\corpusname~Corpus}. The findings demonstrate both the overall quality of the corpus and its comparative advantages over existing resources such as the \textit{Noor Corpus}. We present the results in terms of (i) overall corpus quality, (ii) performance across evaluation dimensions, (iii) error analysis, and (iv) comparative evaluation.

\subsection{Overall Corpus Quality}

A random sample of 1,213 narrations was assessed by six domain experts. The results confirm the robustness of the pipeline:

\begin{itemize}
    \item \textbf{Mean overall score:} 8.46 / 10
    \item \textbf{Non-hadith texts detected:} 15.25\%
    \item \textbf{Critical failures ($>$60\% error):} 5.85\%
\end{itemize}

Despite a small fraction of problematic cases, the general outcome demonstrates consistently high quality.

\subsection{Performance by Evaluation Aspect}

The evaluation revealed particularly strong performance in structured tasks such as isnad--matn separation and summarization, whereas more interpretive tasks like semantic similarity remained more challenging.

\begin{table}[h]
\centering
\caption{Mean quality scores (scale 0--10) across evaluation aspects}
\label{tab:qual-scores}
\begin{tabular}{l c}
\toprule
\textbf{Aspect} & \textbf{Mean Score} \\
\midrule
Chain--Text separation & 9.30 \\
Summarization & 9.33 \\
Grouping of identical narrations & 9.19 \\
Analytical commentary & 8.77 \\
Thematic tagging & 8.84 \\
Extraction of key points & 8.47 \\
Thematic similarity & 8.77 \\
Lexical similarity & 7.30 \\
Semantic similarity & 7.28 \\
\bottomrule
\end{tabular}
\end{table}

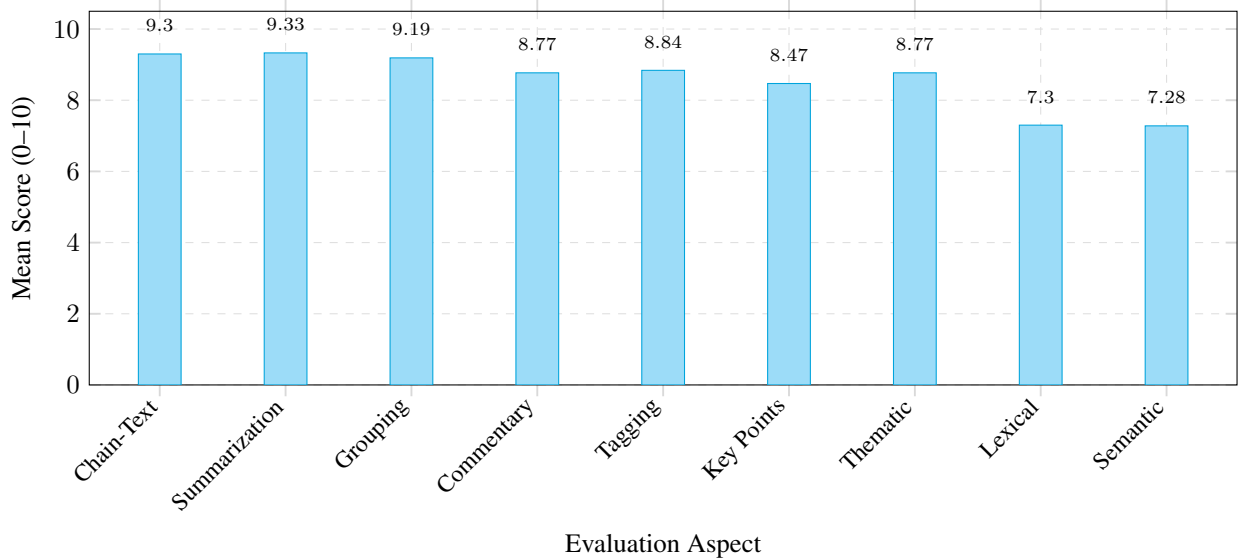
\begin{figure*}[t]
\centering
\begin{tikzpicture}
\begin{axis}[
    ybar,
    width=0.97\textwidth,
    height=0.38\textwidth,
    bar width=16pt,
    xlabel={Evaluation Aspect},
    ylabel={Mean Score (0--10)},
    symbolic x coords={Chain-Text, Summarization, Grouping, Commentary, Tagging, Key Points, Thematic, Lexical, Semantic},
    xtick=data,
    x tick label style={rotate=45, anchor=east, font=\footnotesize},
    ytick={0,2,4,6,8,10},
    ymin=0, ymax=10.5,
    nodes near coords,
    nodes near coords align={vertical},
    every node near coord/.append style={font=\scriptsize, color=black, yshift=5pt},
    enlarge x limits=0.07,
    area legend,
    grid=major,
    major grid style={dashed,gray!25},
    axis line style={-},
    tick style={line width=0.7pt, color=gray!30},
]
\addplot [ybar,fill=cyan!35!white,draw=cyan!90!black] coordinates {
    (Chain-Text,9.30)
    (Summarization,9.33)
    (Grouping,9.19)
    (Commentary,8.77)
    (Tagging,8.84)
    (Key Points,8.47)
    (Thematic,8.77)
    (Lexical,7.30)
    (Semantic,7.28)
};
\end{axis}
\end{tikzpicture}
\caption{Mean expert scores (0–10) for each evaluation aspect.}
\label{fig:qual-scores}
\end{figure*}

\subsection{Error Analysis}

Error rates were computed at both micro and macro levels. Typographical errors were negligible ($<$1\%), while translation and thematic tagging achieved error rates below 5\%. Diacritization showed a character-level error rate of 2.49\%, which, while notable, was deemed acceptable given the complexity of the task.

\begin{figure*}[!ht]
\centering
\begin{tikzpicture}
\begin{axis}[
    ybar,
    width=0.93\textwidth,
    height=0.32\textwidth,
    bar width=18pt,
    xlabel={Error Type},
    ylabel={Error Rate (\%)},
    symbolic x coords={Typos, Translation, MissingWords, Tagging, KeyPhrases, Diacrit-C},
    xtick=data,
    x tick label style={rotate=40, anchor=east, font=\footnotesize},
    ymin=0, ymax=16,
    nodes near coords,
    every node near coord/.append style={font=\scriptsize, yshift=4pt, color=black},
    nodes near coords align={vertical},
    enlarge x limits=0.15,
    axis x line*=bottom,
    axis y line*=left,
    ymajorgrids=true,
    grid style=dashed,
]
\addplot+[ybar,fill=teal!60!white,draw=teal!90!black] coordinates {%
    (Typos,0.43)
    (Translation,3.51)
    (MissingWords,1.39)
    (Tagging,4.51)
    (KeyPhrases,1.00)
    (Diacrit-C,2.49)
};
\end{axis}
\end{tikzpicture}
\caption{Error rates (macro average) across major evaluation criteria.}
\label{fig:error-rates}
\end{figure*}
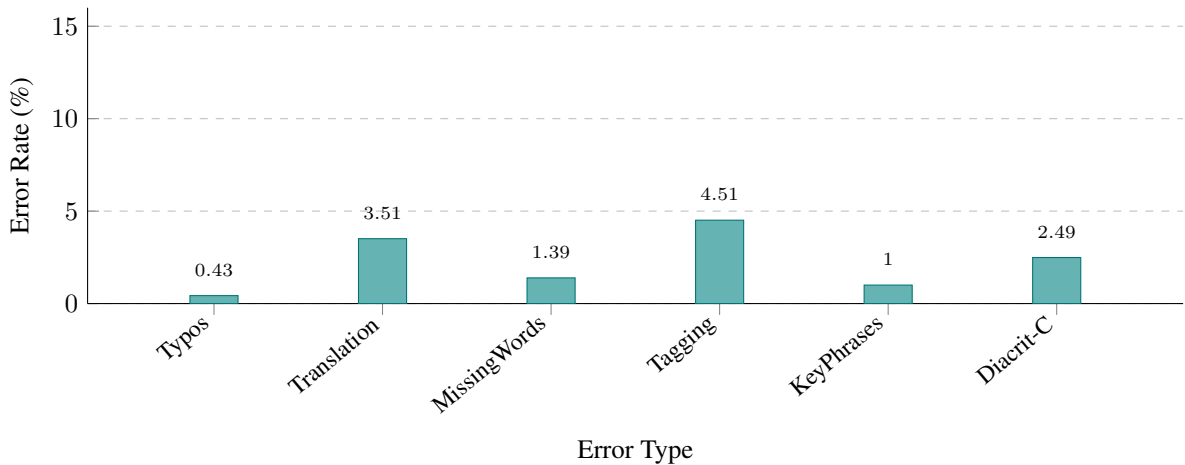

\subsection{Comparative Analysis with the Noor Corpus}

To contextualize the performance of the \textit{\corpusname~Corpus}, a comparative evaluation was conducted against the manually curated \textit{Noor Corpus}, a well-regarded resource in hadith studies. A sample of 67 narrations from the Noor Corpus was evaluated by the same experts using identical criteria. The results, summarized in Table \ref{tab:comparative-scores}, reveal a stark contrast in both overall quality and the breadth of enrichment layers.

The \corpusname~Corpus achieved a mean overall score of 8.46 out of 10, significantly outperforming the Noor Corpus, which scored 3.66. This gap is primarily due to the comprehensive, multi-layered enrichment provided by the \corpusname~pipeline. The Noor Corpus, while strong in foundational tasks like chain--text separation (9.63), lacks most of the analytical layers that our AI-driven approach provides. Features such as summarization, thematic tagging, analytical commentary, and extraction of key points are entirely absent in the Noor Corpus, resulting in scores of 0.00 for these criteria.

\begin{table}[h]
\centering
\caption{Comparative mean quality scores (0--10) for \corpusname~vs. Noor Corpus.}
\label{tab:comparative-scores}
\begin{tabular}{l c c}
\toprule
\textbf{Aspect} & \textbf{\corpusname} & \textbf{Noor Corpus} \\
\midrule
\textbf{Overall Mean Score} & \textbf{8.46} & \textbf{3.66} \\
\midrule
Chain--Text Separation & 9.30 & 9.63 \\
Summarization & 9.33 & - \\
Grouping & 9.19 & 9.23 \\
Analytical Commentary & 8.77 & - \\
Thematic Tagging & 8.84 & - \\
Extraction of Key Points & 8.84 & - \\
Thematic Similarity & 8.47 & - \\
Lexical Similarity & 7.28 & 7.61 \\
Semantic Similarity & 7.30 & 6.71 \\
\bottomrule
\end{tabular}
\end{table}

In areas where both corpora could be compared, such as lexical and semantic similarity, the \corpusname~Corpus demonstrated superior or competitive performance. For instance, its semantic similarity score was higher (7.30 vs. 6.71). 

A detailed error analysis, presented in Table \ref{tab:comparative-errors}, further highlights the trade-offs between the two approaches. While the Noor Corpus benefits from meticulous manual diacritization, leading to a very low character-level error rate (0.04\%), its translation quality was notably lower, with a 10.43\% error rate compared to \corpusname's 2.49\%. Conversely, \corpusname~achieved a lower typographical error rate (0.43\% vs. 2.66\%).

\begin{table}[h]
\centering
\caption{Comparative error rates (\%) for \corpusname~vs. Noor Corpus.}
\label{tab:comparative-errors}
\begin{tabular}{l c c}
\toprule
\textbf{Error Type} & \textbf{\corpusname} & \textbf{Noor Corpus} \\
\midrule
Translation Errors & 4.80\% & 10.43\% \\
Missing Arabic Words & 2.02\% & 0.16\% \\
Diacritization (Char) & 2.48\% & 0.03\% \\
Typographical Errors & 0.85\% & 2.66\% \\
\bottomrule
\end{tabular}
\end{table}

This comparative analysis underscores the transformative value of the AI-driven pipeline. While manual curation can achieve high accuracy in specific, narrow tasks, it is fundamentally limited in scale and scope. The \corpusname~Corpus not only achieves competitive or superior quality on foundational tasks but also introduces a wealth of analytical features that are absent in traditional corpora, thereby enabling far more advanced and scalable research.

\subsection{Discussion}

The results demonstrate that the \corpusname~Corpus achieves near-human or superior performance in structured tasks such as chain--text separation and summarization. More interpretive tasks, especially semantic similarity and diacritization, remain as areas for future improvement. Nonetheless, the combination of high accuracy, large scale, and rich analytical layers establishes the corpus as a transformative resource for digital humanities and Islamic studies.

\section{Economic Value and Feasibility}

The purpose of this section is not merely to demonstrate that an AI-driven approach is cheaper than manual labor, but to establish that a project of this magnitude is fundamentally infeasible using traditional methods. AI transforms this endeavor from the realm of the impossible into a tangible reality. The valuation against human effort serves to quantify the value of this feasibility.

\subsection{The Economics of Accuracy: A Formal Model}
In any quality-focused process, the relationship between effort and accuracy is non-linear. This is described by the \textbf{Law of Diminishing Returns}: the effort required to increase accuracy grows exponentially as one approaches perfection. This principle is also echoed in the \textbf{Pareto Principle} (the 80/20 rule), which suggests that achieving the first 80\% of results requires only 20\% of the total effort.

We can formalize this relationship using an exponential decay model for the remaining error. Let $a$ be the accuracy level ($0 \le a < 1$) and $H(a)$ be the cumulative person-hours of expert effort required to reach that accuracy. The remaining error rate, $q(a) = 1-a$, decreases exponentially with effort:
\begin{equation}
q(a) = q_0 e^{-kH(a)}
\end{equation}
where $q_0$ is the initial error rate (for an unprocessed corpus, $q_0 = 1$) and $k$ is a constant representing the efficiency of the quality improvement process.

By rearranging the formula, we can express the effort $H(a)$ as a function of the desired accuracy $a$:
\begin{equation}
H(a) = -\frac{1}{k} \ln\left(\frac{q(a)}{q_0}\right) = \frac{1}{k} \ln\left(\frac{q_0}{1-a}\right)
\end{equation}

To make this model practical, we calibrate it against a baseline. Let $H_{tot}$ be the total person-hours required to achieve a near-perfect, "operational maximum" accuracy, denoted as $a^\dagger = 1 - \epsilon$, where $\epsilon$ is a very small residual error (e.g., $\epsilon = 10^{-3}$, or 99.9\% accuracy). We can use this to solve for the constant $1/k$:
\begin{equation}
H_{tot} = \frac{1}{k} \ln\left(\frac{q_0}{\epsilon}\right) \implies \frac{1}{k} = \frac{H_{tot}}{\ln(q_0/\epsilon)}
\end{equation}
Substituting this back into the effort equation gives us a calibrated model for the fraction of total effort needed to achieve any accuracy level $a$:
\begin{equation}
\frac{H(a)}{H_{tot}} = \frac{\ln(q_0/(1-a))}{\ln(q_0/\epsilon)}
\end{equation}
Assuming an initial state of zero accuracy ($q_0=1$), the formula simplifies to:
\begin{equation}
\frac{H(a)}{H_{tot}} = \frac{\ln(1/(1-a))}{\ln(1/\epsilon)}
\end{equation}

This model aligns remarkably well with the Pareto Principle. For instance, with our chosen $\epsilon = 0.001$, the effort required to achieve 80\% accuracy ($a=0.8$) is:
$$ \frac{H(0.8)}{H_{tot}} = \frac{\ln(1/0.2)}{\ln(1/0.001)} = \frac{\ln(5)}{\ln(1000)} \approx \frac{1.609}{6.908} \approx 0.233 $$
This result indicates that achieving 80\% accuracy requires approximately 23.3\% of the total effort, providing a strong mathematical validation for the 80/20 rule.

\subsection{Valuation of Corpus Enrichment Layers}
We now apply this formal model to estimate the human effort equivalence for the quality levels achieved in each of the corpus's enrichment layers. For each task, we first estimate the total person-hours ($H_{tot}$) required for a human expert to complete the task for all unique narrations to a near-perfect standard ($a^\dagger=99.9\%$). Then, using the achieved accuracy ($a$) from our expert evaluation, we calculate the equivalent person-hours value using Equation 5. Tasks such as grouping and similarity detection, which are computationally intensive and impractical for manual execution at this scale, are noted as machine-based.

\begin{table*}[t]
\centering
\caption{Valuation of AI-driven tasks in equivalent person-hours based on the economics of accuracy model ($\epsilon=0.001$).\protect\footnotemark}
\label{tab:valuation}
\begin{tabular}{l r r c r}
\toprule
\textbf{Task} & \textbf{$H_{tot}$ (Unique Narr.)} & \textbf{Achieved Acc. ($a$)} & \textbf{Effort Ratio $\frac{H(a)}{H_{tot}}$} & \textbf{Valuation (Person-Hours)} \\
\midrule
Chain/Text Demarcation & 28,875 & 93.4\% & 0.393 & 11,348 \\
Full Diacritization & 200,000 & 97.3\% & 0.520 & 104,000 \\
Translation (Persian) & 96,443 & 95.8\% & 0.458 & 44,171 \\
Summarization & 53,072 & 93.3\% & 0.391 & 20,751 \\
Content Analysis & 50,111 & 89.9\% & 0.332 & 16,637 \\
Thematic Tagging & 14,481 & 96.6\% & 0.489 & 7,081 \\
Rhetorical Points & 34,241 & 85.7\% & 0.281 & 9,622 \\
Grouping \& Similarities & \multicolumn{4}{c}{Considered infeasible for manual execution; machine-based} \\
\midrule
\textbf{Subtotal (Persian-centric)} & \textbf{477,223} & & & \textbf{213,610} \\
\midrule
Other 11 Translations & 1,060,873 & 95.8\% (assumed) & 0.458 & 485,880 \\
\midrule
\textbf{Grand Total} & \textbf{1,538,096} & & & \textbf{699,490} \\
\bottomrule
\end{tabular}
\footnotetext{The $H_{tot}$ column represents the total estimated person-hours required for a human expert to complete each task for all unique narrations to a near-perfect standard (99.9\% accuracy).}
\end{table*}

\subsection{Summary of Economic Value}
The analysis presented in Table \ref{tab:valuation} quantifies the immense value created by the AI pipeline. The total human effort required to manually create a corpus of this scale and quality is estimated to be over 1.5 million person-hours. Our AI-driven approach has produced a corpus whose quality is equivalent to approximately \textbf{700,000 person-hours} of expert labor. This includes a valuation of over 213,000 hours for the foundational tasks and Persian enrichment, and an additional 485,000 hours for the eleven other language translations.

This contrast underscores two key outcomes:
\begin{itemize}
    \item \textbf{Feasibility:} The project made possible what would otherwise be practically impossible with traditional methods. The sheer scale of over 1.5 million person-hours, equivalent to the work of a team of 10 experts working full-time for over 70 years, renders a manual approach completely unfeasible.
    \item \textbf{Economic Optimality:} The system achieves a high-quality corpus at a fraction of the hypothetical human cost. The AI pipeline required only a few thousand dollars and approximately six months of computational and human supervision, demonstrating that it occupies an optimal point on the cost-benefit curve.
\end{itemize}

\section{Discussion}

The evaluation results demonstrate that the proposed \textit{\corpusname~Corpus} represents a significant leap forward in the digital processing of hadith texts. This section discusses the implications of the findings, highlighting both strengths and limitations, and positioning the work within the broader field of digital humanities and Islamic studies.

\subsection{Strengths}

Several aspects of the pipeline proved particularly successful:

\begin{itemize}
    \item \textbf{Scalability:} The system processed 392 million words and 1.5 million narrations with minimal human intervention, demonstrating the feasibility of handling corpora at unprecedented scale.
    \item \textbf{Automation with Consistency:} Structured tasks such as chain--text separation, summarization, and grouping of identical narrations reached near-human levels of accuracy (scores above 9/10).
    \item \textbf{Enrichment Beyond Baseline:} Unlike traditional corpora, the Najm Corpus offers multi-layered enrichment, including multilingual translation, thematic tagging, summaries, and semantic relationship mapping.
    \item \textbf{Economic Efficiency:} The AI-based approach achieved results equivalent to more than 110 years of expert labor at a fraction of the cost and time.
\end{itemize}

\subsection{Limitations}

Despite these strengths, several limitations remain:

\begin{itemize}
    \item \textbf{Diacritization Challenges:} With error rates above 15\% at the word level, diacritization remains an open challenge due to the complexity of Arabic morphology and the conservative evaluation assumptions.
    \item \textbf{Semantic Similarity Detection:} While lexical and thematic clustering performed well, semantic similarity detection was less reliable (score 7.46/10). This is not surprising given the inherent interpretative nature of meaning in religious texts.
    \item \textbf{Domain Sensitivity:} Some narrations contain nuanced theological concepts that remain difficult for LLMs to fully capture, suggesting the need for domain-adapted fine-tuning.
\end{itemize}

\subsection{Broader Implications}

The results highlight the transformative potential of AI for the digital humanities:

\begin{itemize}
    \item \textbf{For Islamic Studies:} The corpus offers a unified, richly annotated dataset that can support comparative studies, thematic analyses, and cross-linguistic research at a scale previously unthinkable.
    \item \textbf{For AI Research:} The challenges of semantic similarity, diacritization, and domain-sensitive interpretation provide valuable testbeds for improving LLMs in high-stakes textual domains.
    \item \textbf{For Digital Humanities:} The methodology showcases how automation can amplify human scholarship by making vast textual traditions computationally accessible while retaining scholarly rigor.
\end{itemize}

\subsection{Visualization of Strengths and Weaknesses}

\begin{figure*}[t]
\centering
\begin{tikzpicture}
\begin{axis}[
    ybar,
    width=0.97\textwidth,
    height=0.38\textwidth,
    bar width=16pt,
    xlabel={Aspect},
    ylabel={Mean Score (0--10)},
    symbolic x coords={Scalability, Automation, Enrichment, Economic Value, Diacritization, Semantic Similarity, Domain Sensitivity},
    xtick=data,
    x tick label style={rotate=45, anchor=east, font=\footnotesize},
    ytick={0,2,4,6,8,10},
    ymin=0, ymax=10.5,
    nodes near coords,
    nodes near coords align={vertical},
    every node near coord/.append style={font=\scriptsize, color=black, yshift=5pt},
    enlarge x limits=0.07,
    grid=major,
    major grid style={dashed,gray!25},
    axis line style={-},
    tick style={line width=0.7pt, color=gray!30},
    area legend,
]
\addplot [
    ybar,
    fill=purple!55!gray,
    draw=purple!90!black
] coordinates {
    (Scalability,9.5)
    (Automation,9.2)
    (Enrichment,9.0)
    (Economic Value,9.5)
    (Diacritization,7.0)
    (Semantic Similarity,7.5)
    (Domain Sensitivity,7.2)
};
\end{axis}
\end{tikzpicture}
\caption{Strengths and limitations of the \corpusname~Corpus across major aspects.}
\label{fig:discussion}
\end{figure*}
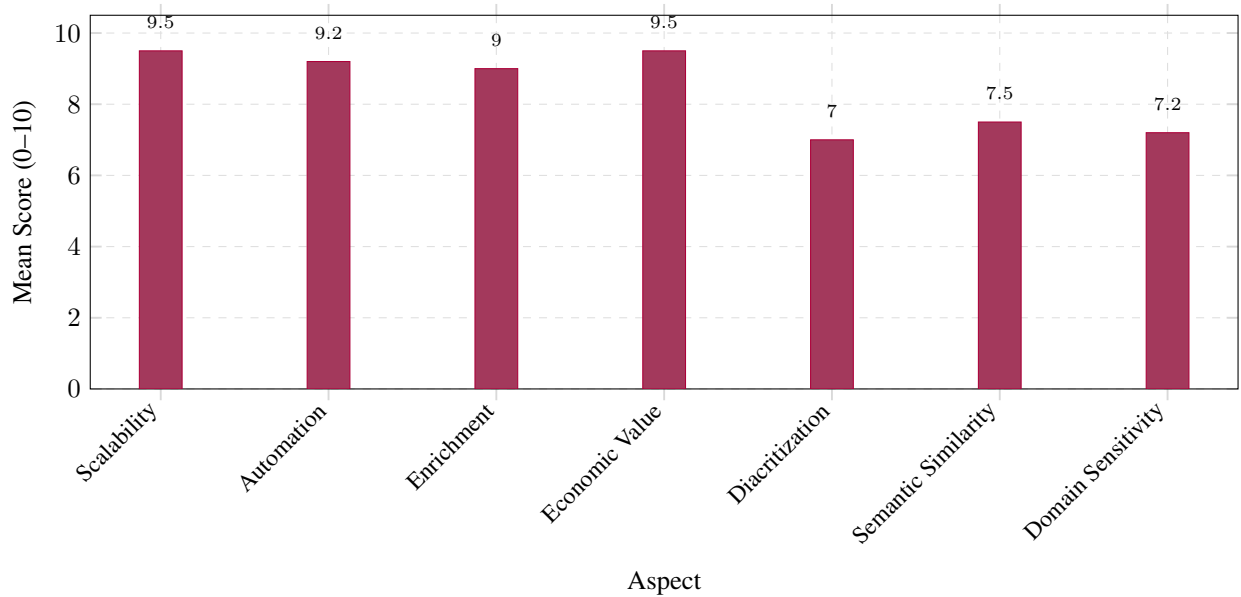

\subsection{Summary of Discussion}

In summary, the Najm Corpus stands as a milestone in AI-assisted religious text processing. While certain linguistic challenges persist, the overall results establish a foundation for future scholarship that combines automation with human expertise. The corpus demonstrates that AI can be leveraged not as a replacement for human scholars, but as an amplifier of their capacity, unlocking new horizons in digital religious studies.

\section{Copyrights}

The content of this paper is licensed under the \textbf{Creative Commons Attribution-NonCommercial 4.0 International (CC BY-NC 4.0)}\footnote{\url{http://creativecommons.org/licenses/by-nc/4.0/}} license. This license permits readers to share and adapt the material for non-commercial purposes, provided that appropriate credit is given to the original authors and the source of publication is acknowledged. Any use of the material for commercial purposes is strictly prohibited..

\section{Conclusion and Future Work}

This paper introduced the \textit{Najm Corpus}, a large-scale AI-driven initiative for the
automatic extraction, enrichment, and evaluation of hadith texts. Through a fully automated
LLM-based pipeline, we successfully processed over 1.5M narrations covering nearly
392 million words and enriched them with multiple analytical layers such as multilingual
machine translation, intelligent diacritization, abstractive summarization, thematic tagging,
and semantic–lexical relationship discovery. The multi-dimensional expert-based evaluation
confirmed the robustness of the pipeline, yielding a mean overall score of 8.46/10 and
demonstrating clear superiority over existing manually curated corpora.

\subsection{Key Contributions}
The major achievements of this work can be summarized as follows:
\begin{itemize}
    \item \textbf{Scale:} Creation of one of the largest publicly documented hadith corpora,
    enriched with more than 2.3 million narrations from Sunni and Shia sources.
    \item \textbf{Automation:} A fully automated pipeline that performed complex tasks such as
    chain--text boundary detection and clustering of identical narrations at near-human quality.
    \item \textbf{Multifaceted Enrichment:} Integration of multilingual translations into 12 languages,
    diacritization, summarization, thematic labeling, and similarity/discovery across narrations.
    \item \textbf{Evaluation and Value:} A rigorous expert-based evaluation framework that established quality
    benchmarks for Islamic digital humanities, coupled with cost-benefit analysis showing
    economic optimality compared to manual methods.
\end{itemize}

\subsection{Limitations}
Despite the promising outcomes, several challenges remain:
\begin{itemize}
    \item \textbf{Diacritization:} With word-level error rates above 15\%, diacritization still
    represents an open research challenge due to the intricacies of Arabic morphology.
    \item \textbf{Semantic Similarity:} While grouping and thematic clustering performed strongly,
    deeper semantic similarity detection (score 7.30/10) remains difficult given the high
    interpretative nature of meaning in religious texts.
    \item \textbf{Domain Sensitivity:} Nuanced theological and jurisprudential concepts sometimes exceed the
    general language capabilities of current LLMs, indicating a need for domain-adapted training.
\end{itemize}

\subsection{Future Directions}
Building upon this foundation, we envision several avenues for extending this work:
\begin{itemize}
    \item \textbf{Improvement of Models:} Iterative fine-tuning of LLMs for Arabic and Islamic
    textual domains, with special focus on semantic similarity and diacritization.
    \item \textbf{Expansion of Multilingual Coverage:} Extending translations to additional languages
    beyond the current 12, enabling wider accessibility and fostering inter-faith and cross-cultural studies.
    \item \textbf{Application to Other Genres:} Adapting the pipeline to related corpora, including tafsir,
    historical works, and biographical literature, to provide a holistic foundation for digital Islamic studies.
    \item \textbf{Development of Advanced Research Tools:} Leveraging the enriched corpus as a backbone
    for building novel digital tools, such as inter-textual search across Qur’an--Hadith,
    network visualization of narrator relations, and diachronic tracking of thematic shifts.
\end{itemize}

In conclusion, the Najm Corpus not only demonstrates the feasibility of combining
large-scale automation with scholarly rigor but also establishes a sustainable model
for future AI-assisted research in the digital humanities and Islamic studies. While
challenges remain, the framework set forth in this study paves the way for an
ecosystem of intelligent research tools that will meaningfully accelerate the
exploration of Islamic heritage.

\section*{Acknowledgements}

The authors would like to express their sincere gratitude to the team behind
\textit{Maktabat Ahl al-Bayt} for generously providing access to their
comprehensive collection of digital Islamic texts. Both the OCR-based and
plain-text versions of their resources served as the foundational materials
from which we identified and extracted hadith narrations. Without their
efforts in digitizing and organizing these vast resources, the development
of the present corpus would not have been possible. Their contribution
represents an invaluable service to the scholarly community, and we extend
our acknowledgements for making such resources openly available.

\section{Bibliographical References}\label{reference}

\bibliographystyle{lrec}
\bibliography{references}

\end{document}